\documentclass[10pt,twocolumn,letterpaper]{article}

\usepackage{cvpr}              

\usepackage{graphicx}
\usepackage{amsmath}
\usepackage{amssymb}
\usepackage{booktabs}
\usepackage{url}
\makeatletter
\@namedef{ver@everyshi.sty}{}
\makeatother
\usepackage{tikz}
\usepackage{comment}
\usepackage{color}
\usepackage{multirow}
\usepackage{gensymb}
\usepackage{caption}
\usepackage{float}
\usepackage{wrapfig}
\usepackage{graphbox}

\usepackage{color, colortbl}

\definecolor{melon}{rgb}{0.99, 0.84, 0.81}

\usepackage[pagebackref,breaklinks,colorlinks]{hyperref}

\usepackage[capitalize]{cleveref}
\crefname{section}{Sec.}{Secs.}
\Crefname{section}{Section}{Sections}
\Crefname{table}{Table}{Tables}
\crefname{table}{Tab.}{Tabs.}

\begin{document}
\title{PointConvFormer: Revenge of the Point-based Convolution}

\author{Wenxuan Wu\textsuperscript{1,2}\thanks{this work was done entirely at Apple Inc., Wenxuan Wu was an intern at Apple Inc. when he participated in the work } , Li Fuxin\textsuperscript{1,3}, Qi Shan\textsuperscript{3}\\
\textsuperscript{1}Oregon State University, \textsuperscript{2}CASIA, \textsuperscript{3}Apple, Inc.\\
{\tt\small \{wuwen, lif\}@oregonstate.edu, \{fli26,qshan\}@apple.com}
}
\maketitle

\begin{abstract}
We introduce PointConvFormer, a novel building block for point cloud based deep network architectures. 
Inspired by generalization theory, PointConvFormer combines ideas from point convolution, where filter weights are only based on relative position, and Transformers which utilize feature-based attention. In PointConvFormer, attention computed from feature difference between points in the neighborhood is used to modify the convolutional weights at each point. Hence, we preserved the invariances from point convolution,  whereas attention helps to select relevant points in the neighborhood for convolution. 
PointConvFormer is suitable for multiple tasks that require details at the point level, such as segmentation and scene flow estimation tasks. 
We experiment on both tasks with multiple datasets including ScanNet, SemanticKitti, FlyingThings3D and KITTI. Our results show that 
PointConvFormer offers a better accuracy-speed tradeoff than classic convolutions, regular transformers, and voxelized sparse convolution approaches. Visualizations show that PointConvFormer performs similarly to convolution on flat areas, whereas the neighborhood selection effect is stronger on object boundaries, showing that it has got the best of both worlds. The code will be available.
\end{abstract}

\section{Introduction}
\label{sec:intro}

\begin{figure}
	\centering
	\includegraphics[width=0.5\textwidth]{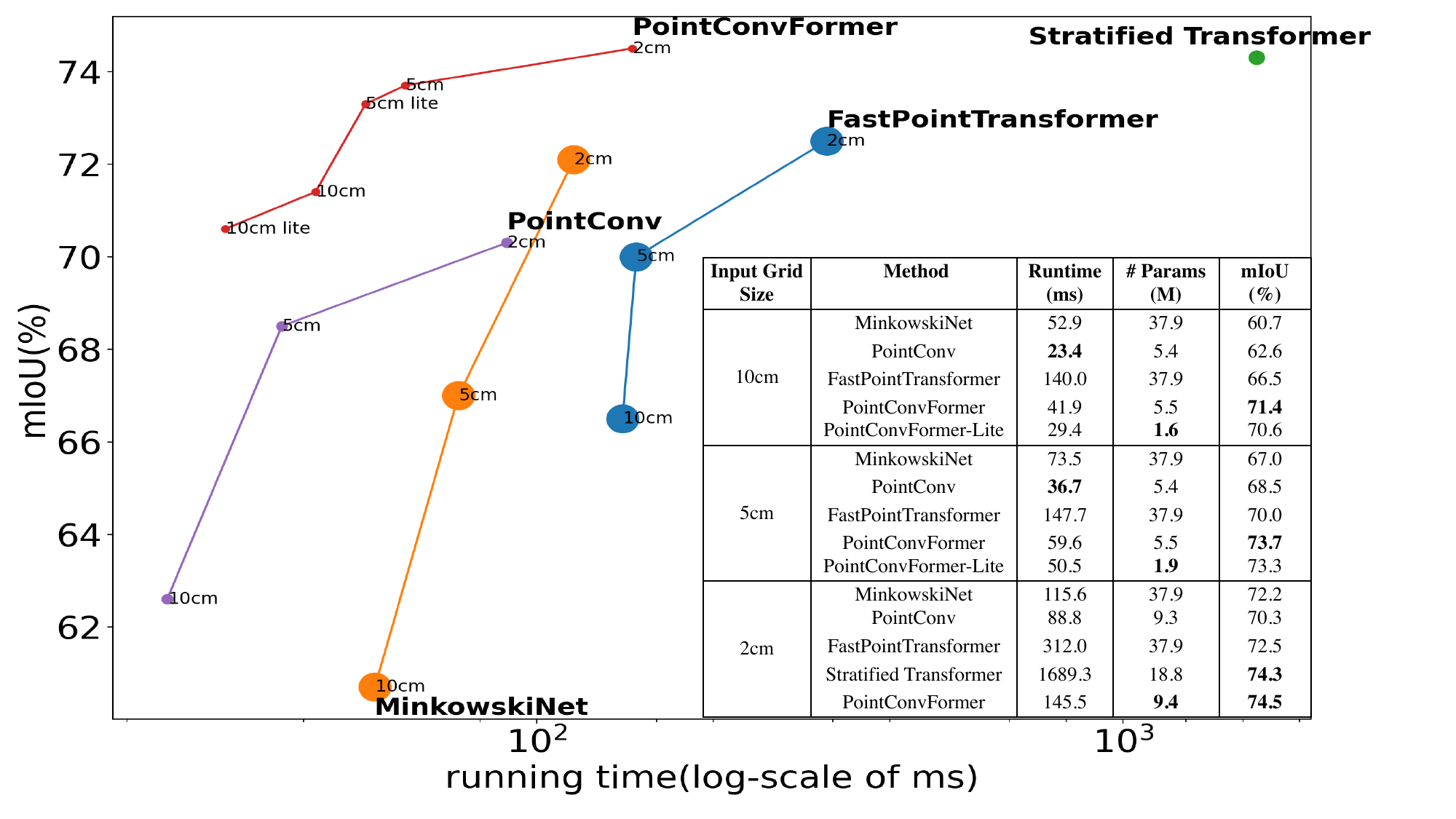}
	\vspace{-0.25in}
	\caption{\small \textbf{Performance vs. running time on ScanNet.} PointConvFormer achieves a  state-of-the-art 74.5\% mIoU while being efficient with faster speed and way less learnable parameters. Larger dot indicates more learnable parameters. All results are reported on a single TITAN RTX GPU }
	\label{fig:mIoU_runtime}
	\vspace{-0.2in}
\end{figure}

Depth sensors for indoor and outdoor 3D scanning have significantly improved in terms of both performance and affordability. Hence, their common data format, the 3D point cloud, has drawn significant attention from academia and industry. Understanding the 3D real world from   point clouds can be applied to many application domains, e.g. robotics, autonomous driving, CAD, and AR/VR. However, unlike image pixels arranged in regular grids, 3D points are unstructured, which makes applying grid based Convolutional Neural Networks (CNNs) difficult. 

Various approaches have been proposed in response to this challenge. ~\cite{su2015multi,li2016vehicle,chen2017multi,kanezaki2018rotationnet,lang2019pointpillars,boulch2017unstructured} introduced interesting ways to project 3D point clouds back to 2D image space and apply 2D convolution. 
Another line of research 
directly voxelizes the 3D space and apply 3D discrete convolution, but it induces massive computation and memory overhead~\cite{maturana2015voxnet,song2017semantic}. Sparse 3D convolution operations~\cite{3DSemanticSegmentationWithSubmanifoldSparseConvNet,choy20194d}  save a significant amount of computation by computing convolution only on  occupied voxels. 

Some approaches directly operate on point clouds~\cite{qi2017pointnet,qi2017pointnet++,su2018splatnet,thomas2019kpconv,wu2019pointconv,li2021devils}. ~\cite{qi2017pointnet,qi2017pointnet++} are pioneers which aggregate information on point clouds using max-pooling layers. Others proposed to reorder the input points with a learned transformation~\cite{li2018pointcnn}, a flexible point kernel~\cite{thomas2019kpconv}, and a convolutional operation that directly work on point clouds~\cite{wang2018deep,wu2019pointconv} which utilizes a multi-layer perceptron (MLP) to learn convolution weights  implicitly as a nonlinear transformation from the relative positions of the local neighbourhood. 

The approach to directly work on points is appealing to us because it allows  direct manipulation of the point coordinates, thus being able to encode rotation/scale invariance/equivariance directly into the convolution weights~\cite{zhang2019rotation,li2021devils,qi2017pointnet++,li2018pointcnn}. These invariances serve as priors to make the models more generalizable. Besides, point-based approaches require less parameters  than voxel-based ones, which need to keep e.g. $3\times 3\times 3$ convolution kernels on all input and output channels. Finally, point-based approaches can utilize k-nearest neighbors (kNN) to find the local neighborhood, thus can adapt to variable sampling densities in different 3D locations. 


\begin{figure*}[t!]
\centering
	\begin{tabular}{cc}
	\includegraphics[align=c,width=0.55\textwidth]{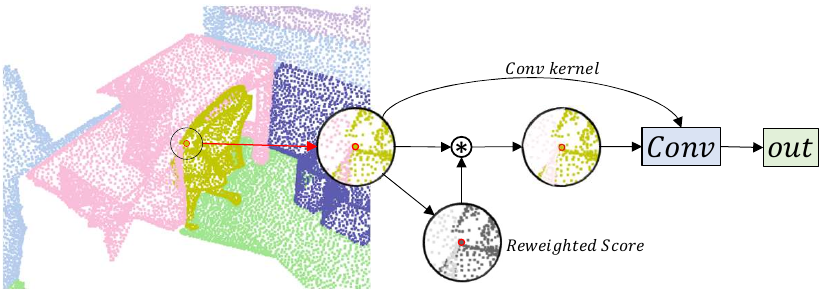}  &
	\includegraphics[align=c,width=0.30\textwidth,height=0.18\textheight]{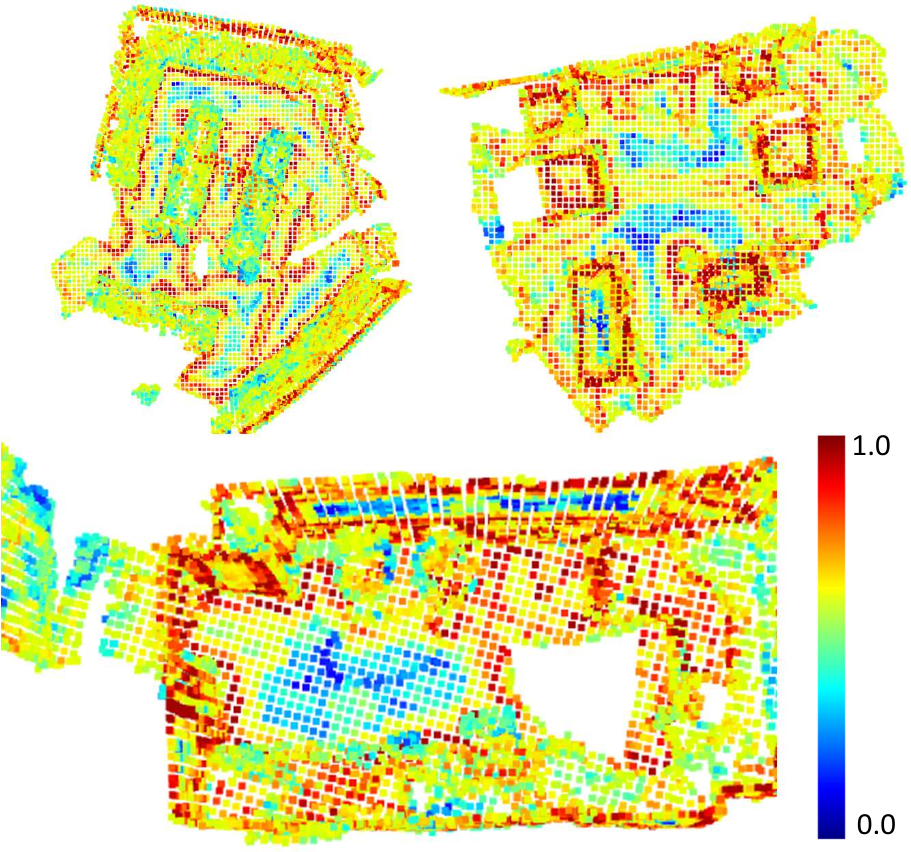} \\
	(a) & (b)
	\end{tabular}
		\vspace{-0.1in}
	\caption{\small (a) \textbf{PointConvFormer}  can be seen as a point convolution, but modulated by a scalar attention  weight for each point in the neighborhood, so that the neighborhood is selectively chosen to perform convolution; (b) Visualization of the reweighting effect in PointConvFormer. The colors are computed by the difference of the maximal attention and the minimal attention in each neighborhood. Red areas have stronger reweighting and blue areas behave similar to convolution. It can be seen that the reweighting effect is stronger at object boundaries where the neighborhoods are more likely to be problematic, whereas on smoother surfaces PointConvFormer behaves more similarly to convolution (more details in Sec.~\ref{sec:visualization})}
	\label{fig:pointconvformer_overview}
	\vskip -0.15in
\end{figure*}

However, so far the methods with the best accuracy-speed tradeoff have still been the sparse voxel-based approaches or a fusion between sparse voxel and point-based models. 
Note that no matter the voxel-based or point-based representation, the information from the input is exactly the same, so it is unclear why fusion is needed. Besides, 
fusion adds significantly to model complexity and memory usage. 
This leads us to question the component that is indeed different between these representations: 
the generalization w.r.t. the irregular local neighbourhood. 
The shape of the kNN neighbourhood in point-based approaches varies in different parts of the point cloud. Irrelevant points from other objects, noise and the background might be included in the neighborhood,  especially around object boundaries, which can be detrimental to the performance and the robustness of point-based models. 

To improve the robustness of models with kNN neighborhoods, we refer back to the generalization theory of CNNs, which indicates that points with significant feature correlation should be included in the same neighborhood~\cite{li2016filter}. A key idea in this paper is that feature correlation can be a way to filter out irrelevant neighbors in a kNN neighborhood, which makes the subsequent convolution more generalizable. We introduce PointConvFormer, which computes attention weights based on feature differences and uses that to reweight the points in the neighborhood in a point-based convolutional model, which indirectly ``improves" the neighborhood for generalization. 

The idea of using feature-based attention is not new, but there are important differences between PointConvFormer and other vision transformers~\cite{dosovitskiy2020image,zhao2021point,park2021fast}. PointConvFormer combines features in the neighborhood with point-wise convolution, whereas Transformer attention models usually adopt softmax attention in this step. 
In our formulation, the positional information is outside the attention, hence 
viewpoint-invariance can be introduced into the convoulutional weights. We believe that invariance helps generalizing across neighborhood (size/rotation) differences between training/testing sets, especially with a kNN neighborhood.

We evaluate PointConvFormer on two point cloud tasks, semantic segmentation and scene flow estimation. For semantic segmentation, experiment results on the indoor  ScanNet~\cite{dai2017scannet}  and the outdoor  SemanticKitti~\cite{behley2019semantickitti}  demonstrate superior performances over classic convolution and transformers with a much more compact network. The performance gap is the most significant at low resolutions, e.g. on ScanNet with a 10cm resolution we achieved more than \textbf{10\% improvement} over MinkowskiNet with only \textbf{15\%} of its parameters (Fig.~\ref{fig:mIoU_runtime}). 
We also apply PointConvFormer as the backbone of PointPWC-Net~\cite{wu2020pointpwc} for scene flow estimation, and observe significant improvements on FlyingThings3D~\cite{MIFDB16} and KITTI scene flow 2015~\cite{menze2018object} datasets as well. These results show that PointConvFormer could potentially compete with sparse convolution as the backbone choice for dense prediction tasks on 3D point clouds. 

\section{Related Work}

\noindent \textbf{Voxel-based networks.} Different from 2D images, 3D point clouds are unordered and scattered in 3D space. A standard approach 
to process  3D point clouds is to voxelize them into regular 3D voxels. However, directly applying dense 3D convolution~\cite{maturana2015voxnet,song2017semantic} onto the 3D voxels can incur massive computation and memory overhead, which limits its applications to large-scale real world scenarios. 
The sparse convolution~\cite{3DSemanticSegmentationWithSubmanifoldSparseConvNet,choy20194d} reduces the convolutional overhead by only working on the non-empty voxels. 
There are also some work~\cite{tang2020searching,tang2022torchsparse} that working on making the network inference more efficient for voxel processing, which could also be extended to point-based methods.

\noindent \textbf{Point-based networks.} 
Point-based  approaches~\cite{qi2017pointnet,qi2017pointnet++,wu2019pointconv,li2021devils,su2018splatnet,wang2018non} directly process point clouds without relying on the voxel structure. ~\cite{qi2017pointnet} propose to use MLPs followed by max-pooling layers to encode and aggregate point cloud features. However, max-pooling can lead to the loss of  critical geometric information in the point cloud. ~\cite{qi2017pointnet++} improves over~\cite{qi2017pointnet} with a hierarchical structure by gradually downsampling and grouping the point cloud with k-nearest neighbourhoods or a query ball method. A number of works~\cite{melekhov2019dgc,jia2016dynamic,mayer2016large,li2019deepgcns,goyal2021kcnet,wang2018local} build a kNN graph from the point cloud and conduct message passing using graph convolution. To better encode the local information, \cite{wang2018deep,xu2018spidercnn,thomas2019kpconv,wu2019pointconv,mao2019interpolated,li2018pointcnn,esteves2018learning,li2021devils} conduct continuous convolution on point clouds. ~\cite{wang2018deep} represents the convolutional weights with MLPs. SpiderCNN~\cite{xu2018spidercnn} uses a family of polynomial functions to approximate the convolution kernels. \cite{su2018splatnet} projects the whole point cloud into a high-dimensional grid for rasterized convolution. ~\cite{wu2019pointconv,thomas2019kpconv} formulate the convolutional weights to be a function of relative position in a local neighbourhood, where the weights can be constructed according to input point clouds. \cite{li2021devils} introduces hand-crafted viewpoint-invariant coordinate transforms on the relative position to increase the robustness of the model. 

\noindent \textbf{Dynamic filters and Transformers.} Recently, the design of dynamic convolutional  filters~\cite{yang2019condconv,zhang2020dynet,chen2020dynamic,jia2016dynamic,wang2019carafe,su2019pixel,zamora2019adaptive,wang2020solov2,tian2020conditional,ma2020weightnet,jampani2016learning,zhou2021decoupled,dai2017deformable,woo2018cbam,cheng20212,boulch2020fkaconv,xu2020squeezesegv3,wu2022point} has drawn more attentions. This line of  work~\cite{ma2020weightnet,zhang2020dynet,chen2020dynamic,yang2019condconv} introduces methods to predict convolutional filters, which are shared across the whole input. \cite{jia2016dynamic,zamora2019adaptive,wang2020solov2,tian2020conditional} propose to predict complete convolutional filters for each pixel. However, their applications  are constrained by their significant runtime and high memory usage. \cite{zhou2021decoupled} introduces decoupled dynamic filters 
 w.r.t. the input features on 2D classification and upsampling tasks. \cite{su2019pixel,tabernik2020spatially} propose to re-weight 2D convolutional kernels with a fixed Gaussian or Gaussian mixture model for pixel-adaptive convolution. Dynamic filtering share some similarities with the popular transformers, whose weights are functions of feature correlations. However, the dynamic filters are mainly designed for images instead of point clouds and focus on regular grid-based convolutions.

With recent success in natural language processing~\cite{devlin2018bert,dai2019transformer,vaswani2017attention,wu2019pay,yang2019xlnet} and 2D images analysis~\cite{hu2019local,dosovitskiy2020image,zhao2020exploring,ramachandran2019stand}, transformers have drawn more attention in the field of 3D scene understanding. Some work~\cite{lee2019set,liu2019point2sequence,yang2019modeling,xie2018attentional} utilize global attention on the whole point cloud. However, these approaches introduce heavy computation overhead and are unable to extend to large scale real world scenes, which usually contain over $100k$ points per point cloud scan. Recently, the work~\cite{zhao2021point,park2021fast,wu2022point} introduce point transformer with local attention to reduce the computation overhead, which could be applied to large scenes. 
Compared to transformers, the attention of the PointConvFormer modulates convolution kernels instead of softmax aggregation.  

\section{PointConvFormer}
\subsection{Point Convolutions and Transformers}

Given a continuous input signal $x(p) \in \mathbb{R}^{c_{in}}$ where $p \in \mathbb{R}^s$ with $s$ being a small number ($2$ for 2D images or $3$ for 3D point clouds, but could be any arbitrary low-dimensional Euclidean space),  $x(\cdot)$ can be sampled as a point cloud $P = \{p_1, \ldots, p_n\}$ with the corresponding values $ x_P = \{x(p_1), \ldots, x(p_n)\}$,  where each $p_i \in \mathbb{R}^s$. The continuous convolution at point $p$ is formulated as: 
\begin{equation}
    Conv(w, x)_p = \int_{\Delta p \in \mathbb{R}^s} \langle w(\Delta p), x(p + \Delta p) \rangle d \Delta p 
\end{equation}
\noindent where $w(\Delta p) \in \mathbb{R}^{c_{in}}$ is the continuous convolution weight function. 
\cite{simonovsky2017dynamic,wu2019pointconv,wang2018non} discretize the continuous convolution on a neighbourhood of point $p$. 
The discretized convolution on point clouds is written as:
\begin{align}
    x'_p &= \sum_{p_i \in \mathcal{N}(p)} w(p_i - p)^\top x(p_i) \label{eq:conv}
\end{align}
\vskip -0.1in
\noindent where $\mathcal{N}(p)$ is a neighborhood that is coonventionally chosen as the $k$-nearest neighbor or $\epsilon$-ball neighborhood of the center point $p$. 
The  function $w(p_i-p): \mathbb{R}^s \mapsto \mathbb{R}^{c_{in}}$ can be approximated as an MLP and learned from data. 
Moreover, because now $p_i - p$ can be explicitly controlled, one can concatenate invariant coordinate transforms on $p_i - p$ as input to $w(\cdot)$, e.g. $\|p_i - p\|$ would be rotation invariant. 
\cite{li2021devils} has found that concatenating a set of rotation and scale-invariant coordinate transforms with $p_i - p$ significantly improves the performance of point convolutions.

In PointConv~\cite{wu2019pointconv}, an efficient formulation was derived when $w(p_i - p)$ has a linear final layer $w(p_i - p)= W_l h(p_i - p)$, where $h(p_i - p): \mathbb{R}^3 \mapsto \mathbb{R}^{c_{mid}}$ is the output of the penultimate layer of the MLP  
and $W_l \in \mathbb{R}^{ c_{in} \times c_{mid}}$ is the learnable parameters in the final linear layer. 
Eq.~(\ref{eq:conv}) can be equivalently written as:
\begin{equation}
\resizebox{0.4\textwidth}{!}{
    $X'_p = \left \langle \mathrm{vec}(W_l), \mathrm{vec}\left\{\sum_{p_i \in \mathcal{N}(p)} h(p_i-p) x(p_i)^\top\right\} \right \rangle.$
    }
    \label{eq:pc_conv}
\end{equation}
where $\mathrm{vec}(\cdot)$ turns the matrix into a vector. Note that $W_l$ represents parameters of a linear layer and hence independent of $p_i$. Thus, when there are $c_{out}$ convolution kernels, $n$ training examples with a neighborhood size of $k$ each, there is no longer a need to store the original convolution weights $w(p_i - p)$ for each point in each neighborhood with a dimensionality of $c_{out} \times c_{in} \times k \times n$. 
Instead, the dimension of all the $h(p_i - p)$ vectors in this case is only $c_{mid} \times k \times n$, where $c_{mid}$ is significantly smaller (usually $4$ to $16$) than $c_{out} \times c_{in}$ (which could go higher than $10^4$). This efficient PointConv enables applications to large-scale  networks on 3D point cloud processing. Besides, its model size is significantly smaller than sparse 3D convolutions, which require $27 \times c_{in} \times c_{out}$ parameters, whereas PointConv only requires $ c_{mid} \times c_{in} \times c_{out}$ for its most costly linear layer $W_l(\cdot)$ ($h(\cdot)$ requires only less than $100$ parameters). As we will see later, $c_{mid}$ can be as small as $4$ for reasonable performance.

Recently, transformer architectures are popular with 2D images. 3D point cloud-based transformers have also been proposed (e.g. \cite{zhao2021point,park2021fast}). Transformers compute an attention model between points (or pixels) based on the features of both points and the positional encodings of them. Relative positional encoding was the most popular which encodes $w(p_i - p)$, similar to Eq.(\ref{eq:conv}). It has been shown to outperform absolute positional encodings~\cite{shaw2018self,chu2021conditional,zhao2021point}. Adopting similar notations to Eq.~\eqref{eq:conv}, we can express the softmax attention model used in transformers as:
\begin{equation}
    \resizebox{0.48\textwidth}{!}{
    $Attention(p) = \sum_{p_i \in \mathcal{N}(p)} \mathrm{softmax}(q(x(p_i)) k(x(p)) + w(p_i - p)) \cdot v(x(p_i))$
    }
    \label{eq:attention}
\end{equation}
where $q(\cdot), k(\cdot), v(\cdot)$ are transformation to the features to form the query, key and value matrices respectively, usually implemented with MLPs. 
Comparing PointConv~\cite{wu2019pointconv} and attention, one can see that both employ $w(p_i - p)$, but in PointConv that is the sole source of the convolutional kernel which is translation-invariant. In attention models, the matching between the query transform $q(x(p_i))$ and the key transform $k(x(p))$ of the features are also considered. 
Besides, having $h(p_i - p)$ outside the attention allows PointConv to utilize an invariant coordinate transform (e.g. scale/rotation) which helps with robustness. 
Transformers usually adopt 2 fully connected layers after the attention module, which are analogous with the $h(\cdot)$ and $W(\cdot)$ layers. Coincidentally, the first fully-connected layer in transformers usually also adopt a dimensionality expansion of $4\times$, which is similar to the minimal $c_{mid}$ we found to be performant in PointConv as well. Hence, an entire transformer block can be seen as similar to a PointConv block, but with the features $x(p)$ also participating in the process of generating weights.

However, one needs to note that in Eq.~(\ref{eq:pc_conv}),  $h(p_i - p)$ creates a set of nonlinear transforms of $p_i - p$ with $c_{mid} > 1$, which does not exist in Eq.~(\ref{eq:attention}) if only a single head is used. This indicates that with Eq.~(\ref{eq:attention}) only nonnegative combinations of the input features can be made. In transformers, this is remedied with the multi-head design where different heads are capable to use different $v(\cdot)$s to enable negative contributions of neighbors. Such a design is not required in PointConv where different $h(\cdot)$ combined with $W_l$ enables both positive and negative contributions of each neighbor.



\subsection{CNN Generalization Theory and The PointConvFormer Layer}
\label{sec:PointConvFormer}

We are interested in adopting the strengths of attention-based models, while still preserve some of the benefits of convolution and explore the possibility of having negative weights. 
To this end, we first look at theoretical insights in terms of which architecture would generalize well. We note the following bound proved in~\cite{li2016filter}:
\begin{equation}
    \hat{G}_N(F) \leq C \max_{p' \in \mathcal{N}(p)} \sqrt{\mathbb{E}_{x,p}[(x(p) - x(p'))^2]} \label{eqn:gaussian_complexity}
\end{equation}
where $\hat{G}_N(F)$ is the empirical Gaussian complexity on the function class $F$: a one-layer CNN followed by a fully-connected layer, and $C$ is a constant. A \textit{smaller} Gaussian complexity leads to better generalization~\cite{bartlett2002rademacher}. To minimize the  bound in Eq.~(\ref{eqn:gaussian_complexity}), one should select points that has high feature correlation to belong to the same neighborhood. In images, nearby pixels usually have the highest color correlation~\cite{li2016filter}, hence conventional CNNs achieve better generalization by choosing a small local neighborhood (e.g. $3\times 3$). Although the bound does not directly apply to transformer models, the idea of using attention to exclude less correlated neighbors could directly improve the generalization bound in Eq.~(\ref{eqn:gaussian_complexity}) for CNNs. 
In 3D point clouds, noisy points can be included in the kNN neighborhood, which 
motivates us to attempt to filter out those noisy points by explicitly checking their $x(p) - x(p')$, keeping only the relevant points in the neighborhood.

Inspired by the discussion above, 
we define a novel convolution operation, \textit{PointConvFormer}, which takes into account both the relative position $p_i - p$ and the feature difference $x(p_i) - x(p)$. The PointConvFormer layer of a point $p$ with its neighbourhood $\mathcal{N}(p)$ can be written as:
\begin{small}
    \begin{align}
    x_p' &= \sum_{p_i \in \mathcal{N}(p)} w(p_i - p)^\top \psi([{x(p_i) - x(p)}, p_i - p]) x(p_i) \label{eq:pointconvformer} 
    \end{align}
\end{small}
\noindent where the function $w(p_i - p)$ 
is the same as defined in Eq.~(\ref{eq:conv}), 
the scalar function $\psi([x(p_i) - x(p), p_i - p]): \mathbb{R}^{c_{in}} \mapsto \mathbb{R}$ is the function of both feature differences $X_{p_i} - X_p$ and position differences. In practice, $\psi(\cdot)$ is approximated with an MLP follwed by an activation layer. 

If we fix the function $\psi(\cdot) = 1$, the PointConvFormer layer is equivalent to Eq.~(\ref{eq:conv}) and reduces to traditional convolution. In Eq.~(\ref{eq:pointconvformer}), 
the function $w(p_i - p)$ learns the weights respect to the relative positions, and the function $\psi(\cdot)$ learns to select useful points in the neighborhood, which works similarly to the attention 
in transformer. However, different from the transformer whose non-negative weights are directly used as a weighted average on the input, the output of $\psi(\cdot)$ only modifies the convolutional filter $w(p_i - p)$, which allows each neighborhood point to have both positive and negative contributions. Hence, our design is more lightweight than ~\cite{wu2022point} in which $\psi$ outputs a vector for each head. 
Besides, 
leaving $w(p_i - p)$ outside $\psi$ allows it to take full advantage of inivariance-aware positional encodings. 

 
Since $\psi(\cdot)$ is outside the convolution, we adopt the same approach as PointConv~\cite{wu2019pointconv} to create an efficient version of the PointConvFormer layer. Following Eq.~(\ref{eq:pc_conv}), we have:
\begin{align}
    x'_p = W_l \sum_{p_i \in \mathcal{N}(p)} h(p_i - p) \psi(x({p_i}), x(p)) x({p_i})^\top \label{eq:pointconvformer2}
\end{align}
\vskip -0.05in
where $W_l$ and $h(\cdot)$ are the same as in Eq.~(\ref{eq:pc_conv}). 

\noindent \textbf{Multi-Head Mechanism}
As in Eq.(\ref{eq:pointconvformer2}), the weight function $\psi(\cdot): \mathbb{R}^{c_{in}} \mapsto \mathbb{R}$ learns the relationship between the center point feature $x(p) \in \mathbb{R}^{c_{in}}$ and its neighbourhood features $x(p_i) \in \mathbb{R}^{c_{in}}$, where $c_{in}$ is the number of the input feature dimension. To increase the representation power of the PointConvFormer, we use the multi-head mechanism to learn different types of neighborhood filtering mechanisms. As a result, the function $\psi: \mathbb{R}^{c_{in}} \mapsto \mathbb{R}$ becomes a set of functions $\psi_i: \mathbb{R}^{c_{in}} \mapsto \mathbb{R}$ with $i \in \{1,...,z\}$, $z$ being the number of heads. 

\subsection{PointConvFormer Block}
\begin{figure}[htb]
	\centering
	\includegraphics[width=0.4\textwidth]{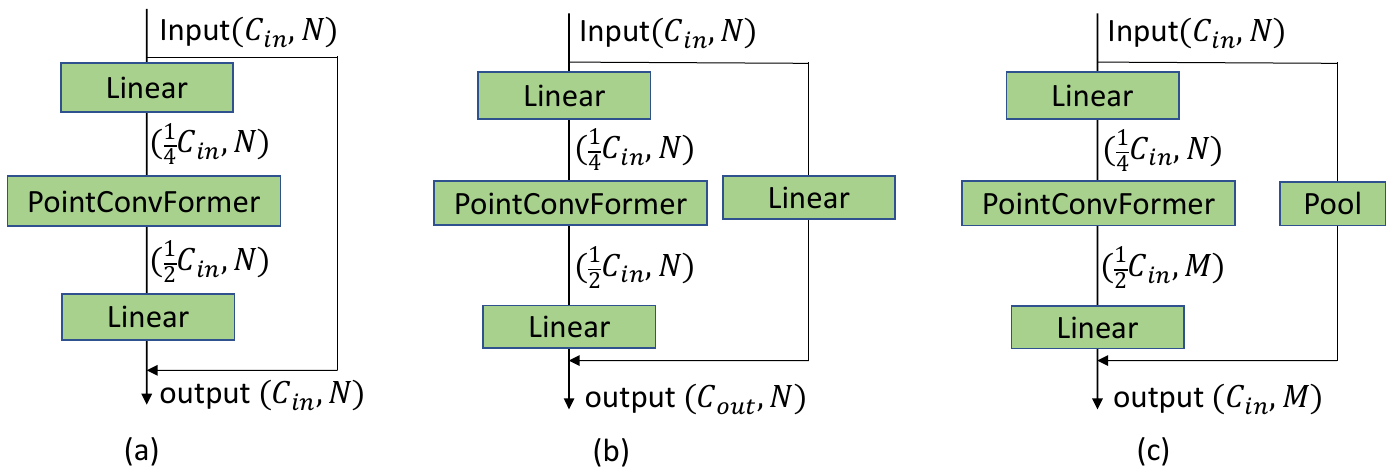}
	\vskip -0.1in
	\caption{\small \textbf{The residual blocks of PointConvFormer.} Number of channels and points are shown in parentheses. The first linear layer reduces the feature dimensionality to $1/4$, then after PointConvFormer, another linear layer upsamples the amount of output channels. Linear and pooling layers are used to change the dimensionality and cardinality of the shortcut as necessary.}
	\label{fig:residual_blocks}
	\vskip -0.15in
\end{figure}

To build deep neural networks for various computer vision tasks, we construct bottleneck residual blocks with PointConvFormer layer as its main components. The detailed structures of the residual blocks are illustrated in Fig.~\ref{fig:residual_blocks}. The input of the residual block is the input point features $X \in \mathbb{R}^{c_{in}}$ along with its  coordinates $p \in \mathbb{R}^3$. The residual block uses a bottleneck structure, which consists of two branches. 
The main branch is a linear layer, followed by PointConvFormer layer, followed by another linear layer. Following ResNet and KPConv~\cite{he2016deep,thomas2019kpconv}, we use one-fourth of the input channels in the first linear layer, conduct PointConvFormer with the smaller number of channels, and finally upsample to the amount of output channels. 
We have found this strategy to significantly reduce the model size and computational cost while maintaining high accuracy for both PointConv and PointConvFormer. 

\noindent \textbf{Downsampling and Deconvolution} We use the grid-subsampling method~\cite{thomas2019kpconv} to downsample the point clouds similar to the $2\times 2$ downsampling in 3D convolutions, which had been shown to outperform random and furthest point downsampling~\cite{thomas2019kpconv}. This subsampling choice makes PointConvFormer comparable with 3D convolution backbones at the same voxelization levels (e.g. 2cm, 5cm) as both use the same voxelization and downsampling strategies. For upsampling layers, we cannot apply PointConvFormer because no feature is available for points that are not part of the downsampled cloud.
Instead, we note that in \eqref{eq:pc_conv} of PointConv, $p$ itself does not have to belong to $\mathcal{N}(p)$, thus we can just apply PointConv layers for deconvolution without features $x(p)$  as long as coordinates $p$ are  known. This helps us to keep the consistency of the network and avoid arbitrary interpolation layers that are not learnable. 

\section{Experiments}

We conduct experiments in a number of domains and tasks to demonstrate the effectiveness of the proposed PointConvFormer in tasks that require point-level accuracy. For 3D semantic segmentation, we use the challenging ScanNet~\cite{dai2017scannet}, a large-scale indoor scene dataset, and the SemanticKitti dataset~\cite{behley2019semantickitti}, a large-scale outdoor scene dataset. Besides, we conduct experiments on the scene flow estimation from 3D point clouds with the synthetic FlyingThings3D dataset~\cite{MIFDB16} for training and the KITTI scene flow 2015 dataset~\cite{menze2018object}. We also conduct ablation studies to explore the properties of the PointConvFormer. 

\noindent \textbf{Implementation Details.} We implement PointConvFormer in PyTorch~\cite{paszke2019pytorch}. The viewpoint-invariant coordinate transform in \cite{li2021devils} is concatenated with the relative coordinates as the input to the $h(\cdot)$ function. We use the AdamW optimizer with $(0.9, 0.999)$ betas and $0.05$  weight decay. For the ScanNet dataset, we train the model with an initial learning rate 0.001 and dropped to half for every 60 epochs for 300 epochs. For the SemanticKitti dataset, the model is trained with an initial learning rate 0.001 and dropped to half for every 8 epochs for 40 epochs. A weighted cross entropy loss is used. To ensure fair comparison with published approaches, we did not employ the recent Mix3D augmentations~\cite{nekrasov2021mix3d} except when it is clearly marked that we used it. For the scene flow estimation, we follow the exact same training pipeline as in~\cite{wu2020pointpwc} for fair comparison.

\subsection{Indoor Scene Semantic Segmentation}

We conduct indoor 3D semantic scene segmentation on the ScanNet~\cite{dai2017scannet} dataset. We use the official split with $1,201$ scenes for training and $312$ for validation. 
We compare against both voxel-based methods such as MinkowskiNet42~\cite{choy20194d} and SparseConvNet~\cite{3DSemanticSegmentationWithSubmanifoldSparseConvNet}, as well as point-based approaches~\cite{qi2017pointnet,wu2019pointconv,li2021devils,yan2020pointasnl,thomas2019kpconv}. Recently, there are work adopting transformer to point clouds. We chose the Point Transformer~\cite{zhao2021point} and the Fast Point Transformer~\cite{park2021fast} as representative transformer based methods. Since the Point Transformer does not report their results on the ScanNet dataset, we adopt their point transformer layer (a standard multi-head attention layer) with the same network structure as ours. Hence, it serves as a direct comparison between PointConvFormer and multi-head attention. 
There exists some other approaches~\cite{chiang2019unified,hu2021bidirectional,hu2021vmnet,kundu2020virtual} which use additional inputs, such as 2D images, which benefit from ImageNet~\cite{deng2009imagenet} pre-training that we do not use. Hence, we excluded these methods from comparison accordingly, but we are comparable to the best of them.

We adopt a general U-Net structure with residual blocks in the encoding layers as our backbone model. 
Through experiments we found out that the decoder can be very lightweight without sacrificing performance (shown in supplementary
). Hence, we set $c_{mid}$ to be $1$ in the decoder throughout the experiments, and just have consecutive PointConv upsampling layers without any residual blocks. Please refer to the supplementary for detailed network structure. Following~\cite{park2021fast}, we conduct experiments on different input voxel sizes, reported in Fig.~\ref{fig:mIoU_runtime} and   Table~\ref{tab:scannet_results_gridsize}. We re-implemented PointConv using the bottleneck architecture which yielded significantly better performance than the original paper, yet still significantly trails our PointConvFormer using the same codebase. 

We compare the top results among work in the literature in Table \ref{tab:scannet_results}. According to Fig.~\ref{fig:mIoU_runtime},  Table.~\ref{tab:scannet_results_gridsize} and Table.~\ref{tab:scannet_results}, our PointConvFormer offers the best accuracy-speed tradeoff regardless of the input grid size. Especially, our PointConvFormer outperforms MinkowskiNet42~\cite{choy20194d} by a significant \textbf{10.1\% with 10cm input grid, 7.3\% with 5cm input grid, and 2.3\% with 2cm input grid}, while being faster than it in the first two cases. It is also significantly \textbf{faster} than all the transformer approaches. On top of this, mix3D could further improve results at 10cm and 5cm resolutions. On 2cm, although mix3D did not improve results on our model with $9.4$M parameters, it helped to propel a much smaller model with $5.8$M parameters to a similar accuracy. 

\begin{figure*}[ht]
	\centering
	\includegraphics[width=0.85\textwidth]{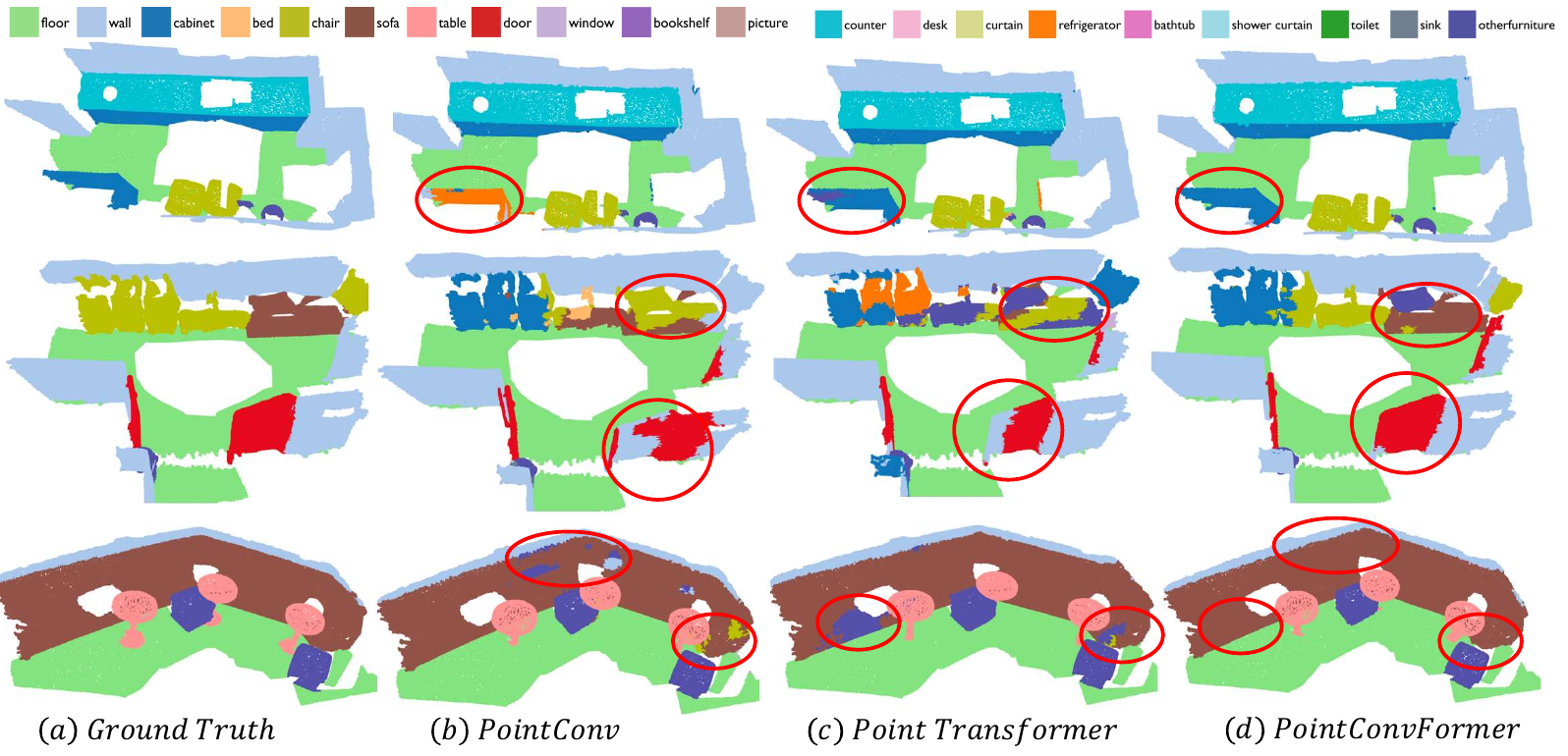}
	\vskip -0.1in
	\caption{\small \textbf{ScanNet result visualization.} We visualize the ScanNet prediction results from our PointConvFormer,  PointConv*~\cite{wu2019pointconv} and Point Transformer~\cite{zhao2021point}. The \textcolor{red}{\textbf{red}} ellipses indicates the improvements of our PointConvFormer over other approaches. Points with ignore labels are filtered for a better visualization.  (Best viewed in color) 
	}
	\label{fig:scannet_visualization}
	\vskip -0.05in
\end{figure*}

We also provide lightweight versions named PointConv-Former-Lite, which utilized less layers in each stage. They are faster and more memory efficient with minimal degradation in performance. 
Fig.~\ref{fig:scannet_visualization} is the visualization of the comparison among PointConv~\cite{wu2019pointconv}, Point Transformer~\cite{zhao2021point} and PointConvFormer on the ScanNet dataset~\cite{dai2017scannet}. We  observe that  PointConvFormer is able to achieve better predictions with fine details comparing with PointConv~\cite{wu2019pointconv} and Point Transformer~\cite{zhao2021point}. Interestingly, it seems that PointConvFormer is usually able to find the better prediction out of PointConv~\cite{wu2019pointconv} and Point Transformer~\cite{zhao2021point}, showing that its novel design brings the best out of both operations.

\begin{table}[htb]
\centering
\caption{\small \textbf{Comparison with different input voxel size.} We compare the results on the ScanNet~\cite{dai2017scannet} validation set with different input voxel size. $^{\dagger}$ means the results are reported in~\cite{park2021fast}. We use grid subsampling~\cite{thomas2019kpconv} to downsample the input point clouds, which is similar to voxelization. However, we still use kNN neighborhood after downsampling which is different from the voxel neighborhood used in other approaches. * means we implemented it under the same codebase and network structure as PointConvFormer}
\vskip -0.1in
\resizebox{0.47\textwidth}{!}{
\begin{tabular}{@{ }l|@{ }c@{ }|@{ }c@{ }cc@{ }}
\hline\noalign{\smallskip}
Methods                   &Voxel/grid size & \# Params(M) & Input  & mIoU(\%)      \\ \noalign{\smallskip}\hline\noalign{\smallskip}
MinkowskiNet42$^{\dagger}$~\cite{choy20194d}& 10cm & 37.9         & Voxel         & 60.5          \\
PointConv* & 10cm & 5.4 & Point & 62.6 \\
Fast Point Transformer~\cite{park2021fast}& 10cm& 37.9         & Voxel                     & 65.9          \\
PointConvFormer(\textbf{ours})      & 10cm & 5.5         & Point                     & \textcolor{blue}{71.4}          \\
PointConvFormer(\textbf{ours}) + mix3D & 10cm & 5.5 & Point & \textbf{72.6} \\
PointConvFormer-Lite (\textbf{ours}) & 10cm & 1.6 & Point & 70.6 \\
\noalign{\smallskip}\hline\noalign{\smallskip}
MinkowskiNet42$^{\dagger}$~\cite{choy20194d}& 5cm& 37.9         & Voxel                     & 66.7          \\
Fast Point Transformer~\cite{park2021fast}& 5cm & 37.9         & Voxel                     & 70.0          \\
PointConv* &  5cm  & 5.4         & Point                     & 68.5          \\
PointConvFormer(\textbf{ours}) & 5cm & 5.5         & Point                     & \textbf{74.0} \\ 
PointConvFormer (\textbf{ours}) + mix3D & 5cm & 5.5 & Point & \textbf{74.3}\\
PointConvFormer-Lite (\textbf{ours}) & 5cm & 1.9 & Point & 73.3 \\
\noalign{\smallskip}\hline\noalign{\smallskip}
MinkowskiNet42$^{\dagger}$~\cite{choy20194d}& 2cm & 37.9         & Voxel         & 71.9          \\
Fast Point Transformer~\cite{park2021fast}& 2cm& 37.9         & Voxel         & 72.1         \\
PointConv* &  2cm  & 5.4         & Point                     & 70.3          \\
PointConvFormer(\textbf{ours})      & 2cm & 9.4         & Point                     & \textbf{74.5}          \\
PointConvFormer (\textbf{ours}) + mix3D & 2cm & 5.8 & Point & \textbf{74.4} \\
PointConvFormer-Lite (ours) & 2cm & 3.8 & Point & 73.3 \\
\noalign{\smallskip}\hline
\end{tabular}%
}
\label{tab:scannet_results_gridsize}
\vskip -0.15in
\end{table}

\begin{table*}[htb]
\centering
\caption{\small \textbf{Semantic segmentation results on ScanNet dataset.} We compare both the ScanNet~\cite{dai2017scannet} validation set and test set. Numbers for baselines are taken from the literature. The numbers for test set are from the official ScanNet benchmark.
} 
\vskip -0.05in
\begin{small}
\begin{tabular}{@{ }l@{ }|@{ }c@{ }|@{ }l@{ }|@{ }c@{ }|@{ }c@{ }|@{ }c@{ }}

\hline\noalign{\smallskip}
Methods                      & \# Params(M) & Input  & Runtime(ms)& Val mIoU(\%)  & Test mIoU(\%)     \\ \noalign{\smallskip}\hline\noalign{\smallskip}
PointNet++~\cite{qi2017pointnet++}& -            & Point & - & 53.5 & 55.7 \\
PointConv~\cite{wu2019pointconv}& -      & Point      &  83.1         & 65.1         & 66.6  \\
KPConv \textit{deform}~\cite{thomas2019kpconv}& 14.9         & Point  &   -        & 69.2       & 68.4  \\
PointASNL~\cite{yan2020pointasnl}& -            & Point     &  -       & 63.5     &  66.6   \\
RandLA-Net~\cite{hu2019randla} & -              & point   & -  & -  &  64.5 \\
VI-PointConv~\cite{li2021devils}& 15.5        & Point     &  88.9    & 70.1        & 67.6 \\ 
SparseConvNet~\cite{3DSemanticSegmentationWithSubmanifoldSparseConvNet}& - & Voxel & - & 69.3   &  72.5 \\
MinkowskiNet42~\cite{choy20194d}   & 37.9         & Voxel     &   115.6       & 72.2        &  73.6 \\
PointTransformer~\cite{zhao2021point} &  -        & Point &   -  & 70.6 &  - \\
Fast Point Transformer~\cite{park2021fast}  & 37.9         & Voxel     &  312.0      & 72.0      &   -  \\
Stratified Transformer~\cite{lai2022stratified}  & 18.8         & point  &  1689.3    & 74.3     &   74.7  \\
PointTransformerV2~\cite{wu2022point}  & 12.8        & point  &  266    & \textbf{75.4}     &   \textbf{75.2}  \\
\noalign{\smallskip}\hline\noalign{\smallskip}
PointConvFormer(ours)         & 9.4         & Point   &  145.5     & 74.5 & \textbf{74.9} \\ \noalign{\smallskip}\hline\noalign{\smallskip}
\end{tabular}%
\end{small}
\label{tab:scannet_results}
\vskip -0.3in
\end{table*}

\subsection{Outdoor Scene Semantic Segmentation}

SemanticKitti~\cite{behley2019semantickitti,geiger2012cvpr} is a large-scale street view point cloud dataset built upon the KITTI Vision Odometry Benchmark~\cite{geiger2012cvpr}. The dataset 
consists of $43,552$ point cloud scans sampled from $22$ driving scene sequences. Each point cloud scan contains $10$ to $13$k points. We follow the training and validation split in~\cite{behley2019semantickitti} and 19 classes are used for training and evaluation. 
For each 3D point, only the $(x,y,z)$ coordinates are given without any color information. It is a challenging dataset because the scanning density is uneven as faraway points are more sparse in LIDAR scans. 

Table~\ref{tab:semantickitti_results} reports the  results on the semanticKitti dataset. Because this work mainly focus on the basic building block, PointConvFormer, which is applicable to any kind of 3D point cloud data, of deep neural network, we did not compare with work~\cite{zhu2021cylindrical,cheng20212} whose main novelties work mostly on LiDAR datasets due to the additional assumption that there are no occlusions from the bird-eye view. 
From the table, one can see that our PointConvFormer outperforms both point-based methods and point+voxel fusion methods. Especially, our method obtains better results comparing with SPVNAS~\cite{tang2020searching}, which utilizes the network architecture search (NAS) techniques and fuses both point and voxel branches. We did not utilize any NAS in our system which would only further improve our performance.

\begin{table}[]
\centering
\caption{\small \textbf{Semantic segmentation results on SemanticKitti validation set.}}
\vskip -0.05in
\begin{small}
\resizebox{0.4\textwidth}{!}{
\begin{tabular}{@{}l@{ }|@{ }c@{ }|@{ }c@{ }|@{ }c@{ }|@{ }c@{ }}
\hline 
Method                &\#MACs& \# Params & Input       & mIoU      \\ 
& (G) & (M) && (\%) \\\hline
RandLA-Net~\cite{hu2020randla}&66.5& 1.2          & Point       & 57.1          \\
FusionNet~\cite{zhang2020deep}& - & -            & Point+Voxel & 63.7          \\
KPRNet~\cite{kochanov2020kprnet} & - & -            & Point+Range & 64.1          \\
MinkowskiNet~\cite{choy20194d}&113.9& 21.7         & Voxel       & 61.1          \\
SPVCNN~\cite{tang2020searching}&118.6& 21.8         & Point+Voxel & 63.8          \\
SPVNAS~\cite{tang2020searching} &64.5& 10.8/12.5    & Point+Voxel & 64.7          \\ \noalign{\smallskip}\hline\noalign{\smallskip}
PointConvFormer &91.1& 8.1          & Point       & \textbf{67.1} \\ \noalign{\smallskip}\hline
\end{tabular}%
}
\end{small}
\label{tab:semantickitti_results}
\vskip -0.15in
\end{table}

\subsection{Scene Flow Estimation from Point Clouds}
Scene flow is the 3D displacement vector between each surface in two consecutive frames. As a fundamental tool for low-level understanding of the world, scene flow can be used in many 3D applications. 
Traditionally, scene flow was estimated directly from RGB data~\cite{huguet2007variational,menze2015object,vogel20153d}. However, with the recent development of  3D sensors such as LiDAR and 3D deep learning techniques, there is increasing interest on directly estimating scene flow from 3D point clouds~\cite{liu2019flownet3d,gu2019hplflownet,wu2020pointpwc,puy2020flot,wei2021pv}. In this work, we adopt PointConvFormer into the PointPWC-Net~\cite{wu2020pointpwc}, which utilizes a coarse-to-fine framework for scene flow estimation, by replacing the PointConv in the feature pyramid layers with the PointConvFormer and keeping the rest of the structure the same as the original version of PointPWC-Net.. 

We name the new network \textit{`PCFPWC-Net'} where PCF stands for PointConvFormer.
To train the PCFPWC-Net we follow the training pipeline in~\cite{wu2020pointpwc}. For a fair comparison, we use the same dataset configurations, hyper-parameters and training pipelines used in~\cite{wu2020pointpwc}. Please check supplementary for more details. 
From Table~\ref{tab:sceneflow_results}, we can see that PCFPWC-Net outperforms previous methods in almost all the evaluation metrics.  Comparing with PointPWC-Net~\cite{wu2020pointpwc}, our PCFPWC-Net achieves around 10\% improvement in all metrics. On the KITTI dataset, our PCFPWC-Net also shows strong result  for scene flow estimation by improving the EPE3D by more than $30\% (0.0694 \mapsto 0.0479)$ over  PointPWC-Net~\cite{wu2020pointpwc}). The qualitative results can be found in supplementary.

\begin{table*}[htb]
\vskip -0.05in
    \centering
    \caption{\small \textbf{Evaluation results on Scene Flow Datasets.} All approaches are trained on FlyingThings3D with the supervised loss. On KITTI, the models are directly evaluated on KITTI without any fine-tuning.}
    \vskip -0.05in
    \begin{small}
    \resizebox{0.8\textwidth}{!}{
    \begin{tabular}{l|l|cccc|cc}
        \hline\noalign{\smallskip}
        Dataset &Method & EPE3D(m)$\downarrow$           & Acc3DS$\uparrow$          & Acc3DR$\uparrow$          & Outliers3D$\downarrow$      & EPE2D(\textit{px})$\downarrow$           & Acc2D$\uparrow$           \\
        \noalign{\smallskip}\hline\noalign{\smallskip}
    \multirow{9}{*}{Flyingthings3D}
        &FlowNet3D~\cite{liu2019flownet3d}    & 0.1136          & 0.4125          & 0.7706          & 0.6016          & 5.9740          & 0.5692          \\
        &HPLFlowNet~\cite{gu2019hplflownet}   & 0.0804          & 0.6144          & 0.8555          & 0.4287          & 4.6723          & 0.6764          \\
        &HCRF-Flow~\cite{li2021hcrf}   & 0.0488          & 0.8337          & 0.9507          & 0.2614          & 2.5652          & 0.8704          \\
        &FLOT~\cite{puy2020flot}   & 0.052          & 0.732          & 0.927          & 0.357          & -          & -          \\
        &PV-RAFT~\cite{wei2021pv}   & 0.0461          & 0.8169          & 0.9574          & 0.2924          & -          & -          \\
        &PointPWC-Net~\cite{wu2020pointpwc}   & 0.0588          & 0.7379          & 0.9276          & 0.3424          & 3.2390          & 0.7994 \\
        &PCFPWC-Net(ours)                  & \textbf{0.0416} & \textbf{0.8645} & \textbf{0.9658} & \textbf{0.2263}  & \textbf{2.2967} & \textbf{0.8871} \\
        \noalign{\smallskip}\hline\noalign{\smallskip}
    \multirow{9}{*}{KITTI}&FlowNet3D~\cite{liu2019flownet3d}   & 0.1767          & 0.3738          & 0.6677          & 0.5271          & 7.2141          & 0.5093          \\
        &HPLFlowNet~\cite{gu2019hplflownet}   & 0.1169 & 0.4783          & 0.7776 & 0.4103          & 4.8055          & 0.5938          \\
        &HCRF-Flow~\cite{li2021hcrf}   & 0.0531          & 0.8631          & 0.9444          & 0.1797          & 2.0700          & 0.8656          \\
        &FLOT~\cite{puy2020flot}   & 0.056          & 0.755          & 0.908          & 0.242          & -          & -          \\
        &PV-RAFT~\cite{wei2021pv}   & 0.0560          & 0.8226          & \textbf{0.9372}          & 0.2163          & -          & -          \\
        &PointPWC-Net~\cite{wu2020pointpwc}   & 0.0694  & 0.7281 & 0.8884 & 0.2648 & 3.0062 & 0.7673 \\
        &PCFPWC-Net(ours)   & \textbf{0.0479} & \textbf{0.8659} & 0.9332 & \textbf{0.1731}  & \textbf{1.7943} & \textbf{0.8924} \\
        \noalign{\smallskip}\hline
    \end{tabular}
    \label{tab:sceneflow_results}
    }    
    \end{small}
    \vskip -0.1in
\end{table*}


\subsection{Visualization of Reweighted Scores}
\label{sec:visualization}
We visualize the difference of the learned attention for a few example scenes in the ScanNet~\cite{dai2017scannet} dataset. The difference is computed by $\max_{x_i \in \mathcal{N}(x_0)}{\psi(x_i)} - \min_{x_i \in \mathcal{N}}\psi(x_i)$, where $\psi$ is the attention score. A larger difference indicates that some points are discarded from the neighborhood. A smaller difference indicates a  nearly constant $\psi$ in the neighbourhood, where PointConvFormer would reduce to regular point convolution. We visualize the difference in Fig.~\ref{fig:pointconvformer_overview}(b).
where it can be seen that larger differences happen mostly in object boundaries. For smooth surfaces and points from the same class, the difference of reweighted scores is low. This visualization further confirms that PointConvFormer is able to utilize feature differences to conduct neighborhood filtering. 

\subsection{Ablation Studies on Attention Types}

In this section, we perform ablation experiment on different attention types in PointConvFormer. The ablation studies are conducted on the ScanNet~\cite{dai2017scannet} dataset with the PointConvFormer-Lite model structure at $5cm$ grid size. Many other ablation studies on different aspects of PointConvFormer are shown in the supplementary. 

We compare the PointConvFormer with three different kind of attention types: Point Transformer~\cite{zhao2021point}, Dot-Product Attention~\cite{vaswani2017attention}, and no attention. The network shares the same backbone architecture, i.e. the same ResNet structure and the same number of layers. 
The dot product version of the attention is shown in this equation:
\begin{equation}
\resizebox{0.4\textwidth}{!}{
    $X_p' = \sum_{p_i \in \mathcal{N}(p)} w(p_i -p)^\top \psi(\frac{1}{\sqrt{d}}\mathbf{q}(X_{p_i})\mathbf{k}(X_p)) X_{p_i}$
    } \label{eq:pcf_qkv}
\end{equation}.

Where $d$ is the dimension of the $q$ and $k$ transforms of the input feature. Theoretically, the computational cost and memory usage of eq. (\ref{eq:pcf_qkv}) are slightly smaller than the  subtractive attention $\psi(X_i - X_j)$ we use in PointConvFormer.

The results in Table \ref{tbl:ablation_vipconv} show that the PointConvFormer attention is significantly better than Point Transformer attention as well as only using the viewpoint-invariant point  convolution~\cite{li2021devils} without attention.
the experiment results in Table~\ref{tbl:ablation_vipconv} also show that the feature difference achieves better results than dot-product attention, which are also confirmed in~\cite{zhao2021point}.  
The dot-product version has a bit more parameters due to the two MLPs for $Q$ and $K$ instead of a single one as in eq.(\ref{eq:pointconvformer}). It in principle uses a bit less memory and computation during inference, however the savings is not very significant in practice due to the small neighborhood size of $K=16$. 
\begin{table}[htb]
\vskip -0.1in
\centering
\caption{\textbf{Different Attention Types.} With the same model architecture and training parameters we change the attention layer of the model. The experiment is performed at the 5cm voxel grid level with the \textit{lite} model architecture}
\vskip -0.1in
\resizebox{0.45\textwidth}{!}{
\begin{tabular}{c|c|c}
\hline
Attention Type & \# Params (M) & mIoU(\%) \\
\hline
PointTransformer Attention  & 2.9 & 71.6 \\
    No Attention (VI-PointConv only) & 1.9 &  71.1 \\
    Dot-Product Attention & 2.1 & 72.0 \\
    PointConvFormer-Lite & 2.0 & \textbf{73.3} \\
    \hline
\end{tabular}
\label{tbl:ablation_vipconv}
}
\vskip -0.1in
\end{table}

\section{Conclusion}

In this work, we propose a novel point cloud layer, PointConvFormer, which can be widely used in various computer vision tasks on point cloud data. Unlike traditional convolution where convolutional kernels are functions of the relative position, the convolutional weights of the PointConvFormer are modified by an attention score computed from feature differences and relative position. Hence, PointConvFormer incorporates benefits of attention models, which could help the network to focus on points with high feature correlation during feature encoding. Experiments on a number of point cloud tasks showed that PointConvFormer is the first point-based model that provides a better accuracy-speed tradeoff w.r.t. sparse 3D convolutional networks.

\subsubsection*{Acknowledgements}
\vskip -0.05in
We thank Dr. Shuangfei Zhai for helpful discussions.

{\small
\bibliographystyle{ieee_fullname}
\bibliography{egbib}
}

\cleardoublepage


\textbf{Supplementary of PointConvFormer: Revenge of the Point-based Convolution}
\setcounter{section}{0}
\section{Network Structure}

\subsection{Network structure for semantic segmentation}

\begin{figure*}
	\centering
	\includegraphics[width=0.95\textwidth]{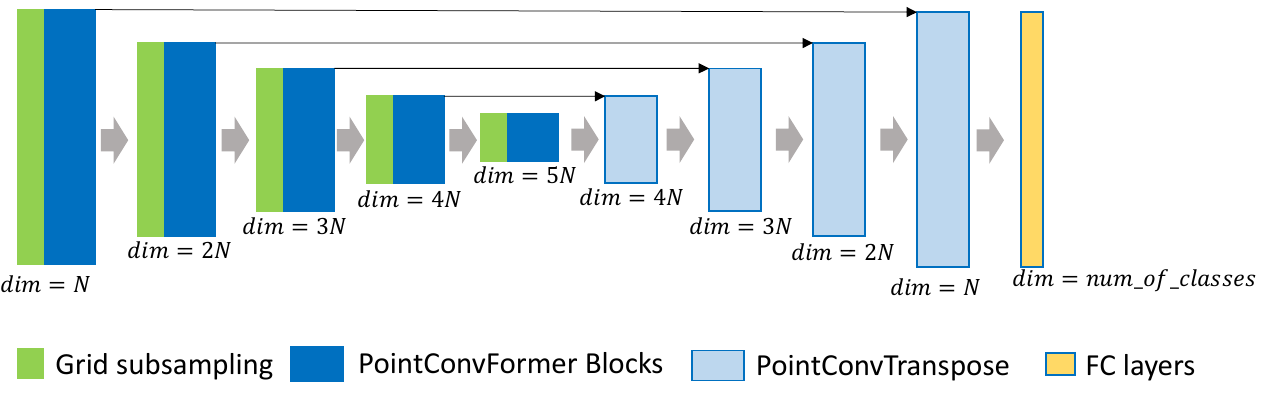}
	\caption{\textbf{The network structure of semantic segmentation.} We use a U-Net structure for semantic segmentation tasks. The U-Net contains 5 resolution levels. For each resolution level, we use grid subsampling to downsample the input point clouds, then followed by several pointconvformer residual blocks. For deconvolution, we just use PointConv as described in the main paper. We set $N = 64$ for ScanNet~\cite{dai2017scannet} Dataset and $N = 48$ for SemanticKitti~\cite{behley2019iccv} Dataset.  (Best viewed in color.)}
	\label{fig:u_net}
\end{figure*}

As in Fig.~\ref{fig:u_net}, we use a U-Net structure for semantic segmentation tasks where we specify a number of base channels, and then have each downsampling stage to use successively more channels. The U-Net contains 5 resolution levels. For each resolution level, we use grid subsampling to downsample the input point clouds, then followed by several pointconvformer residual blocks. The number of layers in the blocks at each level is $[3, 2, 4, 6, 6]$ respectively for the main model. For the \textit{Lite} model at 10cm voxel size, the number of layers is $[3,2,2,2,2]$ at each level, respectively. For the \textit{Lite} model at 5cm voxel size, the number of layers at each level is $[3,3,3,3,3]$, respectively (Table~\ref{tab:num_layers}). For the $9.4$M model at the $2cm$ grid resolution, because it is too fine to be captured by 5 downsampling levels, we utilize a sixth block which contains $2$ layers. For the $5.6$M parameter model at the $2cm$ resolution, we followed PointTransformerv2 to use a set of resolution levels of $[0.02, 0.06, 0.15, 0.375, 0.9375]$, and also replaced the $3$ costly initial layers with a single linear layer, which significantly reduced the size and runtime of the model. This model obtained only $73.0\%$ without mix3D, however, mix3D boosted its performance to $74.4\%$ which is as good as the original model that is significantly larger. For the full model, we have $C_{mid} = 16$, and for the \textit{Lite} model we have $C_{mid} = 4$ which greatly reduced the parameter count with no performance drop at the \textit{Lite} model scale. Latter blocks have more layers since they are cheaper to compute, similar to image convolutional models. For deconvolution, we just use PointConv as described in the main paper. And each block has a single PointConvTranspose layer, which is a PointConv layer that upsamples to locations without any features. For the encoder, we utilize the bottleneck residual architecture described in the paper. For the decoder, because the output dimensionality is always smaller than the input dimensionality, we find that having a bottleneck to $1/4$ of the output dimensionality reduced too much capacity to the model and reduced performance, hence we did not utilize the bottleneck architecture in the decoder and directly performed PointConv on the original input/output dimensionality. 

\begin{table*}[htb]
\centering
    \begin{tabular}{c|c|c|c}
    Input Grid Size     &  2cm & 5cm & 10cm\\
    \hline
    Downsampling levels (cm) & 2, 5, 10, 20, 40, 80 & 5, 10, 20, 40, 80 & 10, 20, 40, 80, 160 \\
    Number of Layers in PointConvFormer & 3, 2,4,6,6,2 & 3, 2,4,6,6 & 3,2,4,6,6 \\
    Number of Layers in PointConvFormer-\textit{Lite} & N/A & 3, 3,3,3,3 & 2,2,2,2,2\\
    \hline
    \end{tabular}
    \caption{Downsampling grid levels and number of layers at each level for different PointConvFormer models}
    \label{tab:num_layers}
\end{table*}

\subsection{Network structure for scene flow estimation}

\begin{figure*}[ht]
	\centering
	\includegraphics[width=0.95\textwidth]{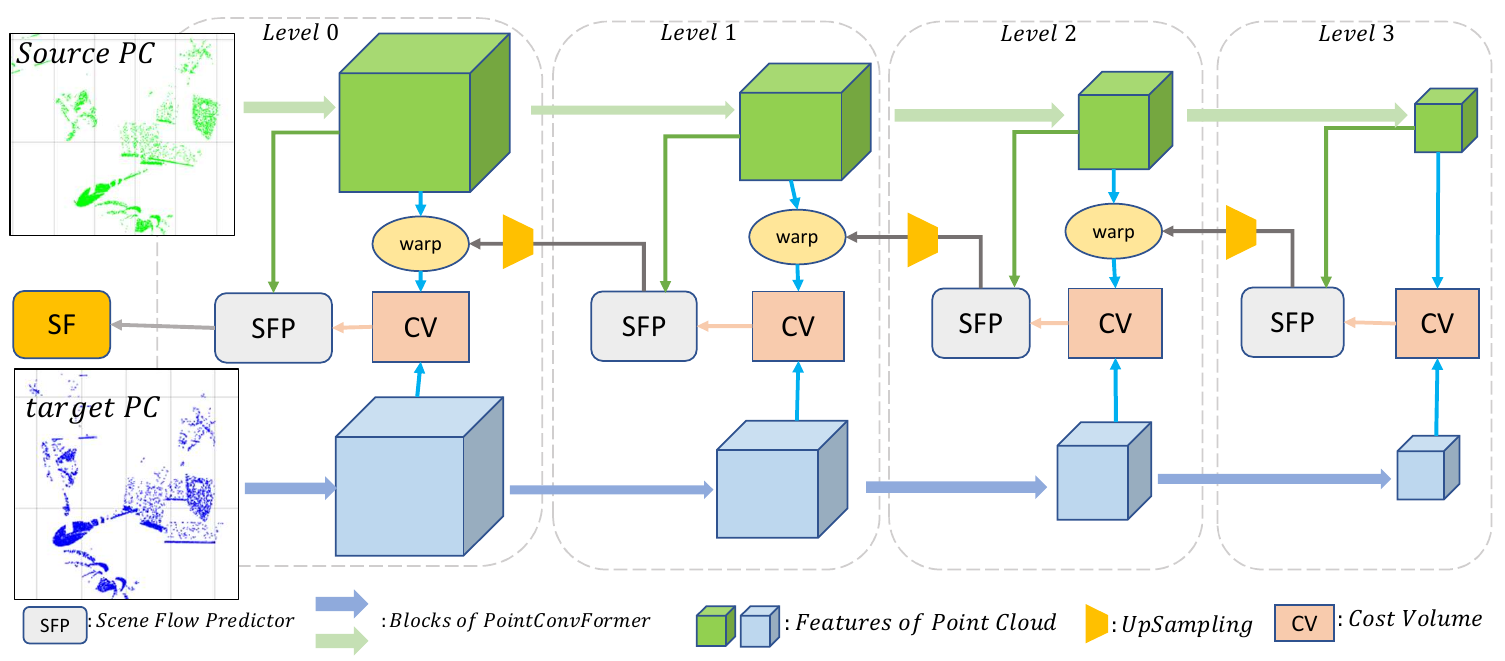}
	\caption{\textbf{The network structure of PointPWC-Net with PointConvFormer.} The feature pyramid is built with blocks of PointConvFormers. As a result, there are 4 resolution levels in the PointPWC-Net. At each level, the features of the source point cloud are warped according to the upsampled coarse flow. Then, the cost volume are computed using the warped source features and target features. Finally, the scene flow predictor predicts finer flow at the current level using a PointConv with features from the first point cloud, the cost volume, and the upsampled flow.  (Best viewed in color.)}
	\label{fig:pointpwc_net}
\end{figure*}

Fig.~\ref{fig:pointpwc_net} illustrates the network structure we used for scene flow estimation.  Following the network structure of PointPWC-Net~\cite{wu2020pointpwc}, which is a coarse-to-fine network design, the PCFPWC-Net also contains 5 modules, including the feature pyramid network, cost volume layers, upsampling layers, warping layers, and the scene flow predictors. We replace the PointConv in the Feature pyramid layers with the PointConvFormer and keep the rest of the structure the same as the original version of PointPWC-Net for fair comparison.

\section{Evaluation Metrics for Scene flow estimation}

\noindent \textbf{Evaluation Metrics.} For comparison, we use the same metrics as~\cite{wu2020pointpwc}. Let $SF_{\Theta}$ denote the predicted scene flow, and $SF_{GT}$ be the ground truth scene flow. The evaluate metrics are computed as follows:

\noindent $\bullet$ \textit{EPE3D(m)}: $\|{SF_{\Theta}-SF_{GT}}\|_2$ averaged over each point in meters.

\noindent $\bullet$ \textit{Acc3DS}: the percentage of points with \textit{EPE3D} $< 0.05m$ or relative error $<5\%$.

\noindent $\bullet$ \textit{Acc3DR}: the percentage of points with \textit{EPE3D} $< 0.1m$ or relative error $<10\%$.

\noindent $\bullet$ \textit{Outliers3D}: the percentage of points with \textit{EPE3D}$>0.3m$ or relative error $>10\%$.

\noindent $\bullet$ \textit{EPE2D(px)}: 2D end point error obtained by projecting point clouds back to the image plane.

\noindent $\bullet$ \textit{Acc2D}: the percentage of points whose \textit{EPE2D} $< 3px$ or relative error $<5\%$.

\section{Ablation Studies}
 

In this section, we perform thorough ablation experiments to investigate our proposed PointConvFormer. The ablation studies are conducted on the ScanNet~\cite{dai2017scannet} dataset. For efficiency, we downsample the input point clouds with a grid-subsampling method~\cite{thomas2019kpconv} with a grid size of $10cm$ as in~\cite{park2021fast}.

\noindent \textbf{Number of neighbours.} We first conduct experiments on the neighbourhood size $k$ in the PointConvFormer for feature aggregation. The results are reported in Table.~\ref{tab:ablation_neighbours}. The best result is achieved with a neighbourhood size of $16$. Larger neighbourhood sizes of $32, 48$ do not introduce significant gains on the result, and $48$ actually decreased the performance a bit, which may be caused by introducing excessive less relevant features in the neighbourhood~\cite{zhao2021point}. We choose $16$ based on similar performance to $32$ and significantly smaller memory footprint and faster speed.

\begin{table}[htb]
\vskip -0.1in
\noindent \begin{minipage}[c]{0.5\textwidth}
\centering
\captionof{table}{\small \textbf{Ablation Study.} Number of neighbours in each local neighbourhood.}
\resizebox{\textwidth}{!}{%
\begin{tabular}{l|ccccc}
\hline
Nieghbourhood Size   & 4     & 8     & 16             & 32    & 48    \\ \hline
mIoU(\%)             & 64.61 & 69.54 & \textbf{71.40} & 71.19 & 69.84 \\ \hline
\end{tabular}%
}
\label{tab:ablation_neighbours}
\end{minipage}
\hspace{0.02\textwidth}
\begin{minipage}[c]{0.45\textwidth}
\centering
\captionof{table}{\small \textbf{Ablation Study.} Number of heads.}
\resizebox{\textwidth}{!}{%
\begin{tabular}{l|cccc}
\hline
Number of Heads & 2     & 4     & 8    &  16 \\ \hline
mIoU(\%)       & 70.71 & 70.58 & \textbf{71.40} &  70.97 \\ \hline
\end{tabular}%
}
\label{tab:ablation_headnumber}
\end{minipage}
\vskip -0.1in
\end{table}

\noindent \textbf{Number of heads in $\psi$.} As described in the main paper, our PointConvFormer could employ the multi-head mechanism to further improve the representation capabilities of the model. We conduct ablation experiments on the number of heads in the PointConvFormer. The results are shown in Table.~\ref{tab:ablation_headnumber}. From Table.~\ref{tab:ablation_headnumber}, we find that PointConvFormer achieves the best result with 8 heads. 

\noindent \textbf{Decoder $c_{mid}$} PointConv and PointConvFormer implementations lead to a dimensionality expansion of the network that is $c_{mid}$ times the size of the input dimensionality, hence significantly increase the number of parameters in the subsequent linear layer $W_l$. One empirical contribution in semantic segmentation we made is that we found that the decoder does not really need this dimensionality expansion, which leads to significant savings in the number of parameters. In Table \ref{tbl:ablation_decoder_cmid}, it is shown that adding $3$ million parameters by using a $c_{mid}$ of $16$ in the decoder only leads to a small improvement of $0.4\%$, hence in our model we choose to set $c_{mid} = 1$ in the decoder of segmentation models, since those parameters could be used better elsewhere. Parameter savings here and the bottleneck blocks allow us to use more layers yet still have a smaller model than \cite{li2021devils}.

\begin{table}[htb]
    \centering
    \begin{tabular}{|c|c|c|c|c|c|}
    \hline
        $c_{mid}$ in decoder & 1 & 3 & 4 & 8 & 16 \\
        \hline
         mIoU (\%) & 71.4 & 71.5 & 70.8 & 71.5 & \textbf{71.8} \\
         \# Params (M) & \textbf{5.48} & 5.90 & 6.11 & 6.96 &  8.64 \\
         \hline
    \end{tabular}
    \caption{Different $c_{mid}$ in the decoder. $c_{mid}$ of 1 in the decoder did not significantly lower the performance, yet saves a significant amount of parameters, hence we choose it in the final model}
    \label{tbl:ablation_decoder_cmid}
\end{table}

\section{More Result Visualizations}

\begin{figure*}[ht]
	\centering
	\includegraphics[width=0.95\textwidth]{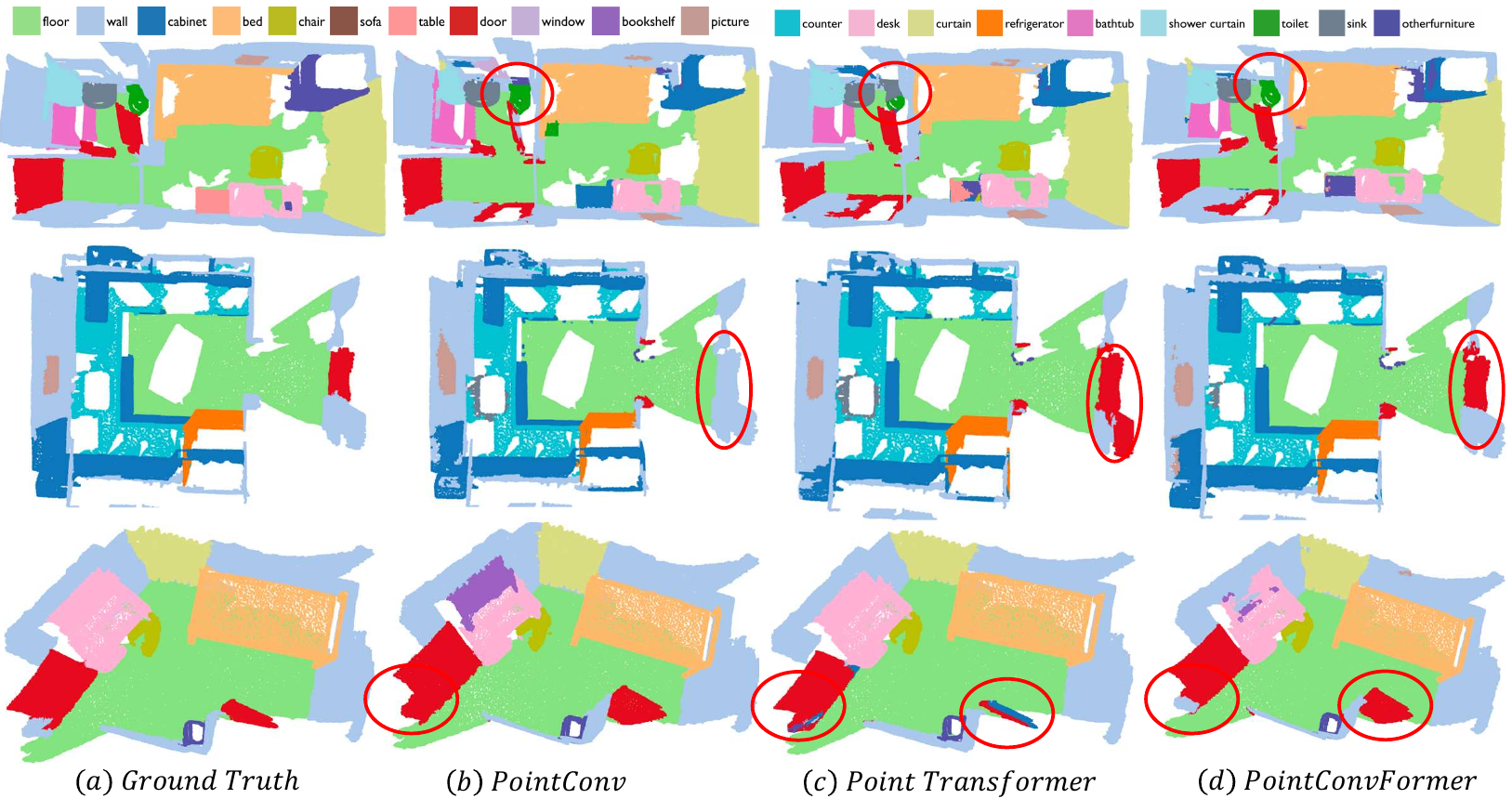}
	\caption{\textbf{ScanNet result visualization.} We visualize the ScanNet prediction results from our PointConvFormer, PointConv~\cite{wu2019pointconv} and Point Transformer~\cite{zhao2021point}. The \textcolor{red}{\textbf{red}} ellipses indicates the improvements of our PointConvFormer over other approaches. Points with ignore labels are filtered for a better visualization.  (Best viewed in color) 
	}
	\label{fig:scannet_visualization_supp}
\end{figure*}

Fig.~\ref{fig:scannet_visualization_supp} is the visualizations of the comparison among PointConv~\cite{wu2019pointconv}, Point Transformer~\cite{zhao2021point} and PointConvFormer on the ScanNet dataset~\cite{dai2017scannet}. We  observe that  PointConvFormer is able to achieve better predictions with fine details comparing with PointConv~\cite{wu2019pointconv} and Point Transformer~\cite{zhao2021point}. Interestingly, it seems that PointConvFormer is usually able to find the better prediction out of PointConv~\cite{wu2019pointconv} and Point Transformer~\cite{zhao2021point}, showing that its novel design brings the best out of both operations. 

\begin{figure*}[t]
	\centering
	\includegraphics[width=0.95\textwidth]{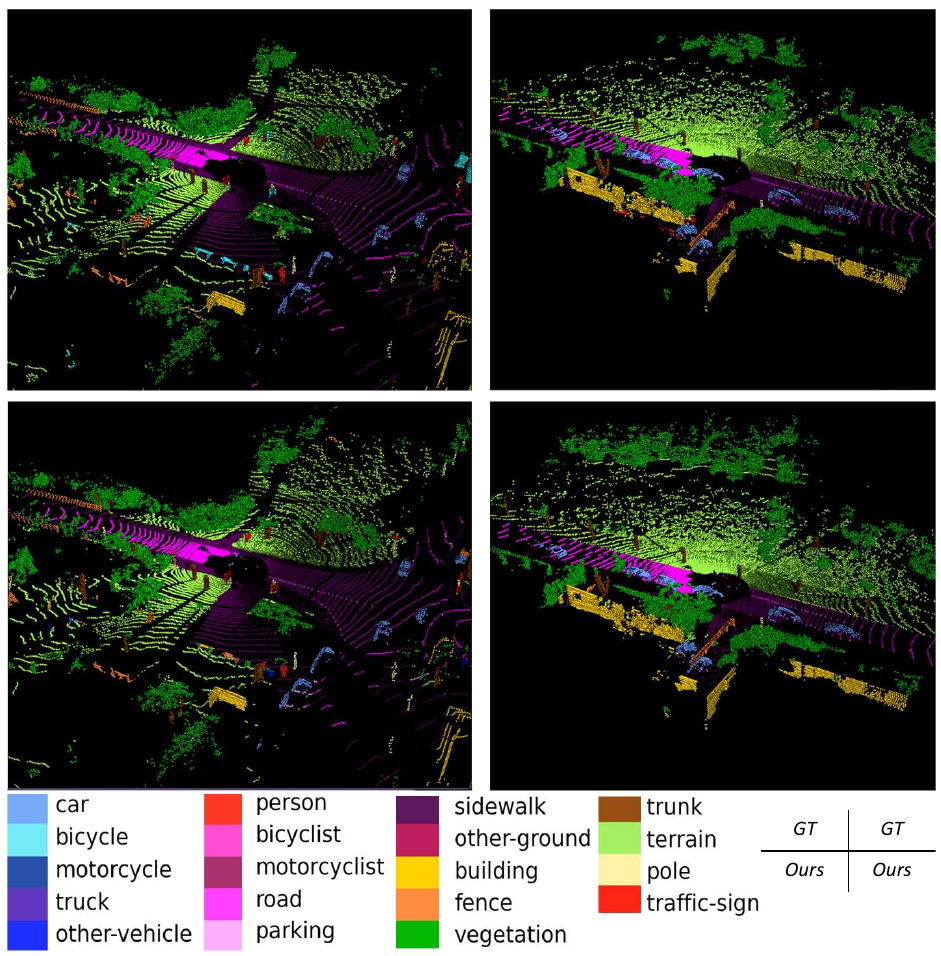}
	\vskip -0.1in
	\caption{\small \textbf{SemanticKitti result visualization.} We visualize the SemanticKitti prediction results from our PointConvFormer. Each column is a scan from SemanticKitti validation set. The first row is the input, the second row is the ground truth, the third row is our prediction. (Best viewed in color.)}
	\label{fig:semantickitti_visualization}
\end{figure*}

\begin{figure*}[t]
	\centering
	\includegraphics[width=0.95\textwidth]{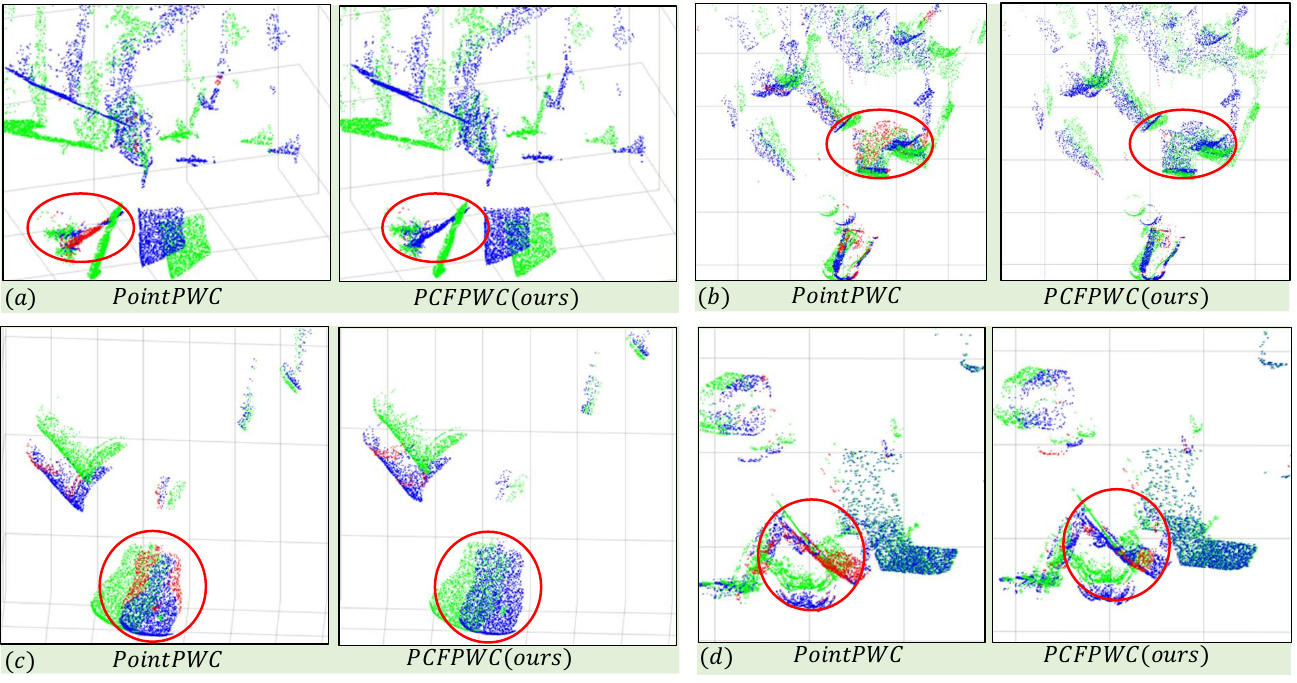}
	\vskip -0.1in
	\caption{\small \textbf{Qualitative comparison between PointPWC-Net and PCFPWC-Net~(FlyingThings3D~\cite{mayer2016large}).} (a) is the visualization of the FlyingThings3D dataset. (b) is the visualization of the KITTI dataset. Green points are the source point cloud. Blue points are the points warped by the correctly predicted scene flow. The predicted scene flow belonging to Acc3DR is regarded as a correct prediction. For the points with incorrect predictions, we use the ground truth scene flow to warp them and the warped results are shown as red points. (Best viewed in color.)}
	\label{fig:sceneflow_visualization_ft3d}
\end{figure*}

\begin{figure*}[t]
	\centering
	\includegraphics[width=0.95\textwidth]{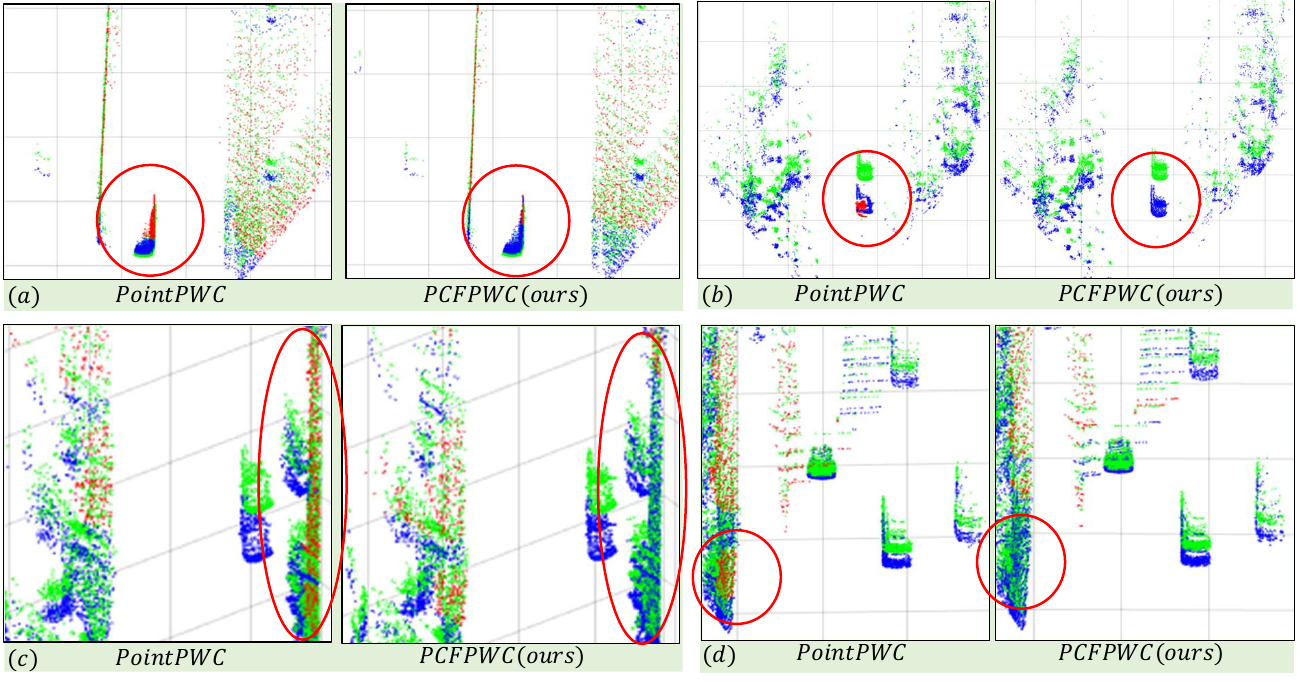}
	\vskip -0.1in
	\caption{\small \textbf{Qualitative comparison between PointPWC-Net and PCFPWC-Net~(KITTI~\cite{menze2015joint}).} Green points are the source point cloud. Blue points are the points warped by the correctly predicted scene flow. The predicted scene flow belonging to Acc3DR is regarded as a correct prediction. For the points with incorrect predictions, we use the ground truth scene flow to warp them and the warped results are shown as red points. (d) is a failure case, where the points on the wall or ground/road are hard to find accurate correspondences for both PointPWC and PCFPWC. (Best viewed in color.)}
	\label{fig:sceneflow_visualization_kitti}
\end{figure*}

Fig.~\ref{fig:semantickitti_visualization} illustrates the prediction of PointConvFormer on the SemanticKitti dataset~\cite{behley2019iccv}. Fig.~\ref{fig:sceneflow_visualization_ft3d}, and Fig.\ref{fig:sceneflow_visualization_kitti} are the comparison between the prediction of PointPWC~\cite{wu2020pointpwc} and PCFPWC-Net on the FlyingThings3D~\cite{MIFDB16} and the KITTI Scene Flow 2015 dataset~\cite{menze2015joint}. Please also refer to the video for better visualization.

\end{document}